\documentclass[lettersize,journal]{IEEEtran}
\usepackage{amsmath,amsfonts}
\usepackage{algorithmic}
\usepackage{array}
\usepackage[caption=false,font=normalsize,labelfont=sf,textfont=sf]{subfig}
\usepackage{textcomp}
\usepackage{stfloats}
\usepackage{url}
\usepackage{verbatim}
\usepackage{graphicx}
\usepackage{enumitem}
\usepackage{amssymb}% to access $\blacksquare$
\usepackage{cite}
\usepackage{xcolor}
\newcounter{boxlblcounter}  % Creating a custom list using the list environment
\def\BibTeX{{\rm B\kern-.05em{\sc i\kern-.025em b}\kern-.08em
    T\kern-.1667em\lower.7ex\hbox{E}\kern-.125emX}}
\usepackage{balance}
\newcommand\Title{SRPT vs Smith Predictor for Vehicle Teleoperation}

\newcommand{\RN}[1]{%
  \textup{\uppercase\expandafter{\romannumeral#1}}%
}

\begin{document}

% Include statement for preprint on the first page
\begin{center}
\textbf{This work has been submitted to the IEEE for possible publication. Copyright may be transferred without notice, after which this version may no longer be accessible.}
\end{center}
% Begin main text on the second page
\newpage

\title{\Title}
%\author{Jai~Prakash \IEEEmembership{Fellow, IEEE},~Michele~Vignati,~Edoardo~Sabbioni,~and~Federico~Cheli% <-this % stops a space
%\thanks{The authors belong to the Department of Mechanical Engineering, Politecnico Di Milano, Italy (e-mail:
%jai.prakash@polimi.it; michele.vignati@polimi.it; edoardo.sabbioni@polimi.it; federico.cheli@polimi.it)
%}%
%}

%\author{Jai~Prakash\,\orcidlink{0000-0001-6660-8615},
%~Michele~Vignati\,\orcidlink{0000-0002-8403-355X},
%~and~Edoardo~Sabbioni\,\orcidlink{0000-0002-4356-8814}% <-this % stops a space
\author{Jai~Prakash\,
~Michele~Vignati\,
~and~Edoardo~Sabbioni % <-this % stops a space
\thanks{The authors belong to the Department of Mechanical Engineering, Politecnico Di Milano, Italy (e-mail:
jai.prakash@polimi.it; michele.vignati@polimi.it; edoardo.sabbioni@polimi.it)
}%
}
%\markboth{IEEE Transactions on Vehicular Technology,~Vol.~XX, No.~X, April~2023}%
{\Title}

\maketitle
\begin{abstract}
Vehicle teleoperation has potential applications in fallback solutions for autonomous vehicles, remote delivery services, and hazardous operations. However, network delays and limited situational awareness can compromise teleoperation performance and increase the cognitive workload of human operators. To address these issues, we previously introduced the novel successive reference pose tracking (SRPT) approach, which transmits successive reference poses to the vehicle instead of steering commands. This paper compares the stability and performance of SRPT with Smith predictor-based approaches for direct vehicle teleoperation in challenging scenarios. The Smith predictor approach is further categorized, one with Lookahead driver and second with Stanley driver. Simulations are conducted in a Simulink environment, considering variable network delays (250–350 ms) and different vehicle speeds (14–26 km/h), and include maneuvers such as tight corners, slalom, low-adhesion roads, and strong crosswinds. The results show that the SRPT approach significantly improves stability and reference tracking performance, with negligible effect of network delays on path tracking. Our findings demonstrate the effectiveness of SRPT in eliminating the detrimental effect of network delays in vehicle teleoperation.

%\textcolor{blue}{Even with no GPS correction, SRPT performs well as EKF pose estimation is negligible in a small time-window.}
\end{abstract}

\begin{IEEEkeywords}
Vehicle teleoperation, remote driving, network delay, latency, time-delay, SRPT, NMPC, Simulink, Smith predictor, Wireless network communication.
\end{IEEEkeywords}

\section*{Nomenclature}
\addcontentsline{toc}{section}{Nomenclature}
\begin{IEEEdescription}[\IEEEusemathlabelsep\IEEEsetlabelwidth{$V_1,V_2,V_3$}]
\item[NMPC] Nonlinear model predictive control.
\item[SRPT] Successive reference pose tracking.
\item[AD] Autonomous driving.
\item[ODD] Operational Design Domain.
\item[FWD] Front wheel drive.
\item[IMU] Inertial measurement unit.
\item[$\tau$] round trip network delay.
\item[$\tau_1$] Uplink delay part of the round trip delay.
\item[$\tau_2$] Downlink delay part of the round trip delay.
\item[$k_1,k_2$] Constants for the Lookahead driver model.
\item[$k$] Constant for the Stanley driver model.
\item[$V_x$] Vehicle longitudinal speed.
\item[$\Delta y$] Cross-track error.
\item[$\delta$] Steer angle.
\item[$\psi$] Vehicle heading angle.
\item[$X_{Ref}$] Reference pose.
\item[$CG$] Center of gravity of the vehicle.
\item[$l_F$] longitudinal distance between front axle and CG.
\item[$L$] Vehicle wheelbase.
\item[$R$] Instantaneous radius of curvature.
\item[$s$] Distance along track length.
\item[$L_{ind}$] Lookahead distance for reference-pose driver model.
\item[${P}^{A}_{B}$] Relative pose of $A$ with respect to $B$.
\item[$\Delta t_{Horizon}$] Time horizon for NMPC prediction.
\item[$\mu$] Road adherence coefficient.
\end{IEEEdescription}
\section{Introduction}

\begin{figure}[h]
\centering
    \includegraphics[width=1.0\columnwidth]{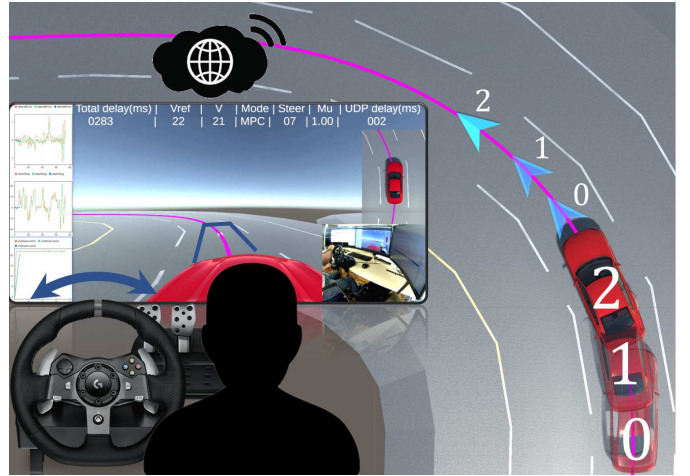}
    \caption{A pictorial representation of SRPT approach for direct vehicle teleoperation. The remote vehicle receives successive reference poses as it moves forward.}
    \label{fig:00_abstractImgPPT}%
\end{figure}

\IEEEPARstart{A}{utomated} vehicles (AVs) have garnered increasing attention as a potential solution for future mobility. However, the deployment of AVs is still hindered by various difficulties and edge cases that have yet to be fully resolved. Teleoperation has emerged as a backup plan for AVs, offering a way to remotely support an AV when it reaches the limits of its operational design domain. Teleoperation is the remote control of a device or a vehicle from a distance. This can be done using either wired communication or wireless communication. Here the vehicle is a mobile robot that can be controlled remotely, typically wirelessly.
The use of teleoperation technology is to offer a secure and effective method to get over these restrictions anytime an AD function hits the limits of its ODD. The AV can resume its voyage in full automation after it has been returned to its nominal ODD \cite{Domagoj2022}. Vehicle teleoperation has the potential to also revolutionize various industries, such as autonomous taxi service, industrial equipment teleoperation, disaster response, and military operations.

Despite having great potential, vehicle teleoperation is currently facing various challenges, such as problems with human-machine interaction, limited situational awareness, network latency, and control loop instability. Although the challenges are significant, we are primarily focusing on reducing the detrimental impact of network latency. By doing so, we are aiming to improve the stability of the control loop, which in turn will enhance the safety and effectiveness of teleoperation systems, and ultimately help to achieve more reliable and efficient teleoperation of vehicles.

The time delays in vehicle teleoperation tasks reduce the accuracy and speed at which human operators can perform a remote task \cite{7139813, Luck2006}. Significant delays can cause overcorrection by the operator, resulting in oscillations that impair teleoperation performance and may even destabilize the control loop \cite{Sheridan1993, Gorsich2018EvaluatingMP}. Over time, various concepts for teleoperating road vehicles have been developed and researched\cite{Domagoj2022}, including direct control, shared control, trajectory guidance, interactive path planning, and perception modification. Direct control \cite{Georg2020, Mutzenich2021, Hoffmann2021, chucholowski2016, Tang2014, Argelaguet2020, Georg2019, Georg2020_2} involves the operator viewing sensor data and sending control signals like steering and throttle, but it suffers from reduced situational awareness and transmission latency. Shared control \cite{Saparia2021, Anderson2013, Schimpe2020, Qiao2021, Schitz2021, Justin2017} has a shared controller inside the vehicle that assesses operator commands to avoid collisions, improving safety but still suffering from latency. Trajectory guidance \cite{Gnatzig2012, Hoffmann2022, Kay1995, Bjornberg1504690, Schitz2021_2} involves the vehicle following a path and speed profile generated by the operator without being affected by network latency, although real-time profile generation is unfeasible. Interactive path planning \cite{Hosseini2014, Schitz2021_3} uses the vehicle's perception module to calculate optimal paths, which the operator confirms to follow, bypassing network latency but requiring a functional set of AD perception module. Perception modification \cite{Feiler2021} involves the operator identifying false-positive obstacles to support the AD perception module, which largely depends on the availability of AD perception module.

Except for the direct control concept, all other vehicle teleoperation concepts rely on the automated modules and perception of autonomous vehicles. In cases where the perception module fails, the vehicle teleoperation becomes impossible with other teleoperation concepts. Therefore, it is essential to ensure the independence of vehicle teleoperation from the perception module, making it a fallback option for autonomous vehicles. This work aims to strengthen the direct control concept of vehicle teleoperation to support this notion.

\subsection{Related Work}
The use of predictive displays has been found to be effective in compensating for delays and improving vehicle mobility in human-in-the-loop experiments \cite{Chucholowski, 7504430, Brudnak2016PredictiveDF, jai2022, tito2015, Zheng2019, Zheng2020}. This approach considers the delay and control signals from the operator to estimate the vehicle position and displays it to the operator as a ``third-person view" \cite{tito2015}. Predictive models can be either model-based \cite{jai2022}, model-free \cite{Zheng2019}, or a combination of both \cite{Zheng2020}. Model-based predictors require a vehicle model to predict the vehicle's response, and prediction accuracy depends on the accuracy of the vehicle model. However, in the presence of unknown disturbances in driving scenarios, such as low adhesion road or crosswind, the prediction accuracy deteriorates. On the other hand, model-free approaches make predictions by considering the delayed state dynamics received from the vehicle but suffer from convergence time in state prediction, resulting in delayed prediction. Combining both approaches results in improved operation but not significantly. In summary, predictive displays aim to bypass the time delay in loops by predicting states using delayed states as input. This is useful for human-in-loop teleoperation, as it allows the human operator to control the vehicle in real-time without waiting for feedback. However, if the prediction accuracy decreases, the chances of asynchrony increase.

\subsection{Previous Work}
In our previous research \cite{jai2022_2}, we introduced the pose-based control strategy, SRPT, for vehicle teleoperation. The control station transmits the reference pose, which is the intended vehicle pose, to the vehicle discretely at a rate of 30 Hz. At the control station, the driver model, which is a reference-pose decider, considers the received (delayed) vehicle pose from the remote vehicle and the prior known mission plan. On the remote vehicle side, the vehicle controller optimizes steer and speed commands while considering actuator constraints and environmental disturbances to reach the reference pose. To sense environmental disturbances, SRPT utilizes IMU sensors installed in the vehicle. We previously evaluated SRPT \cite{jai2022_3}, which combines the benefits of direct control and waypoint guidance concepts. In that paper, the human operator created the waypoints (the reference poses) by steering the lateral position of the augmented lookahead vehicle (blue) outline on the display using joystick steering (figure \ref{fig:00_abstractImgPPT}).

\subsection{Contribution of Paper}

This paper focuses on assessing the performance improvement of the SRPT approach for vehicle teleoperation by comparing it with the Smith prediction strategy. In the Smith prediction strategy, after predicting vehicle states, two types of driver models are assessed. One is the Lookahead driver model and second is the Stanley driver model. The test track consists of maneuvers with progressively increasing difficulty. The experiments are performed in a Simulink simulation environment, where variable network delays (250-350ms) and a 14-dof vehicle model for the main vehicle are considered.

Overall, our experiments show that the SRPT approach outperforms the Smith prediction strategy in terms of accuracy, and stability, especially for challenging maneuvers. These findings demonstrate the potential of the SRPT approach to improve the safety and efficiency of vehicle teleoperation in real-world applications.

\subsection{Outline of Paper}
The rest of the paper is organized as follows. Section \ref{sec2.1} presents the characteristics of network delay. Section \ref{sec2.2} presents the Smith predictor with two driver models. Section \ref{sec2.3} explains the SRPT mode. Section \ref{sec3} provides an overview of the simulation platform. Section \ref{sec4} discusses the experimental structure. Section \ref{sec5} presents and discusses the results. Section \ref{sec6} concludes with the work summary, key findings, and future work.
\section{Method}\label{sec2}
\subsection{Network delays}\label{sec2.1}
\begin{figure}[h]
\centering
    \includegraphics[width=0.9\columnwidth]{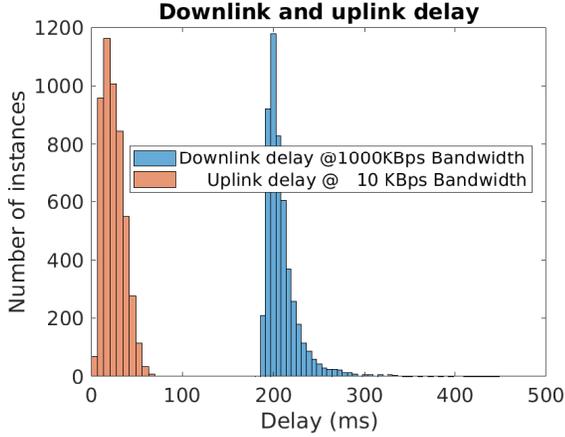}
    \caption{Delays observed in data transmission over 4G \cite{jai2022_2}.}
    \label{fig: delays}%
\end{figure}
The time delay involved in vehicle teleoperation from the perspective of the control station can be divided into two parts. The first part is known as downlink delay ($\tau_2$), which refers to the time it takes for the streamed images to be received by the control station. The second part is known as uplink delay ($\tau_1$), which refers to the time delay between generating driving commands at the control station and the actuation of those commands at the vehicle. The downlink delay is a lumped sum of various factors, such as camera exposure delay, image encoding time, network delay in transmitting the images, and image-decoding time. The uplink delay, on the other hand, is a sum of network delay in transmitting the driving commands towards the vehicle and the vehicle actuation delay. In case of wireless communication using 4G, variability is associated with both downlink and uplink delays. The corresponding delays with utilized bandwidth are presented in figure (\ref{fig: delays}), where 5000 picture frames and driving commands are considered in a typical urban setting, with the vehicle connected to 4G mobile connectivity and the control station connected via wired LAN to the internet. The $\tau_1$ is measured at the vehicle by subtracting the timestamp of driving commands from the current timestamp, and $\tau_2$ is measured at the control station by subtracting the timestamp of an image received from the current timestamp.

\subsection{Smith predictor with two types of driver model}\label{sec2.2}

\begin{figure}[h]
\centering
    \includegraphics[width=\columnwidth]{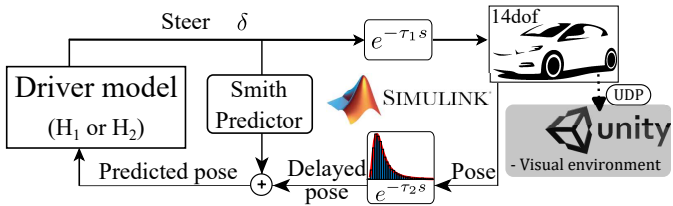}
    \caption{Smith predictor schematic for vehicle teleoperation simulation. H$_1$ and H$_2$ are types of driver models considered. Unity has no role in simulation, it is just to display the manoeuvres.}
    \label{fig:smith_1}%
\end{figure}

The Smith predictor approach \cite{smith} is a popular predictive control method used in bilateral teleoperation. It was first introduced by O.J. Smith in 1957 and is a model-based prediction approach. Figure (\ref{fig:smith_1}) shows a schematic of the Smith predictor in the control loop of vehicle teleoperation systems with variable time delays. The steering input is passed through the Smith predictor block, which outputs a correction term which needs to be added to the (received) delayed pose to predict the current pose of the vehicle. Smith predictor block is further elaborated in our previous work \cite{jai2022_2}, where its transfer function is presented. It provides the human operator with the sense of controlling the vehicle in real-time by predicting the current position of the vehicle, bypassing the network delay. Thereupon, the human operator can steer based on vehicle current pose and the mission plan. In this paper, two types of driver model are considered instead of human volunteers for the sake of reproducibility of results and as a preliminary comparison of the SRPT approach with the Smith predictor approach. %Christoph Popp et al. \cite{Popp2023} suggest that geometry base lateral controller for a vehicle works well for low lateral acceleration scenarios and ideal reference point is located $5–16\%$ of the wheelbase behind the front axle.
Christoph Popp et al. \cite{Popp2023} suggest that geometry base lateral controller for a vehicle works well for low lateral acceleration scenarios.
Considering low-medium speed vehicle teleoperation, below mentioned two driver models are adopted.

\subsubsection{Lookahead driver, H$_1$}\label{sec2.1.1}
\begin{figure}[]
\centering
\subfloat[]{\includegraphics[width=0.3\columnwidth]{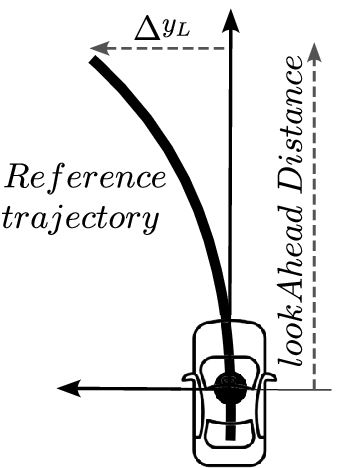}%
\label{fig:03_1lookAhead}}
\hfil
\subfloat[]{\includegraphics[width=0.55\columnwidth]{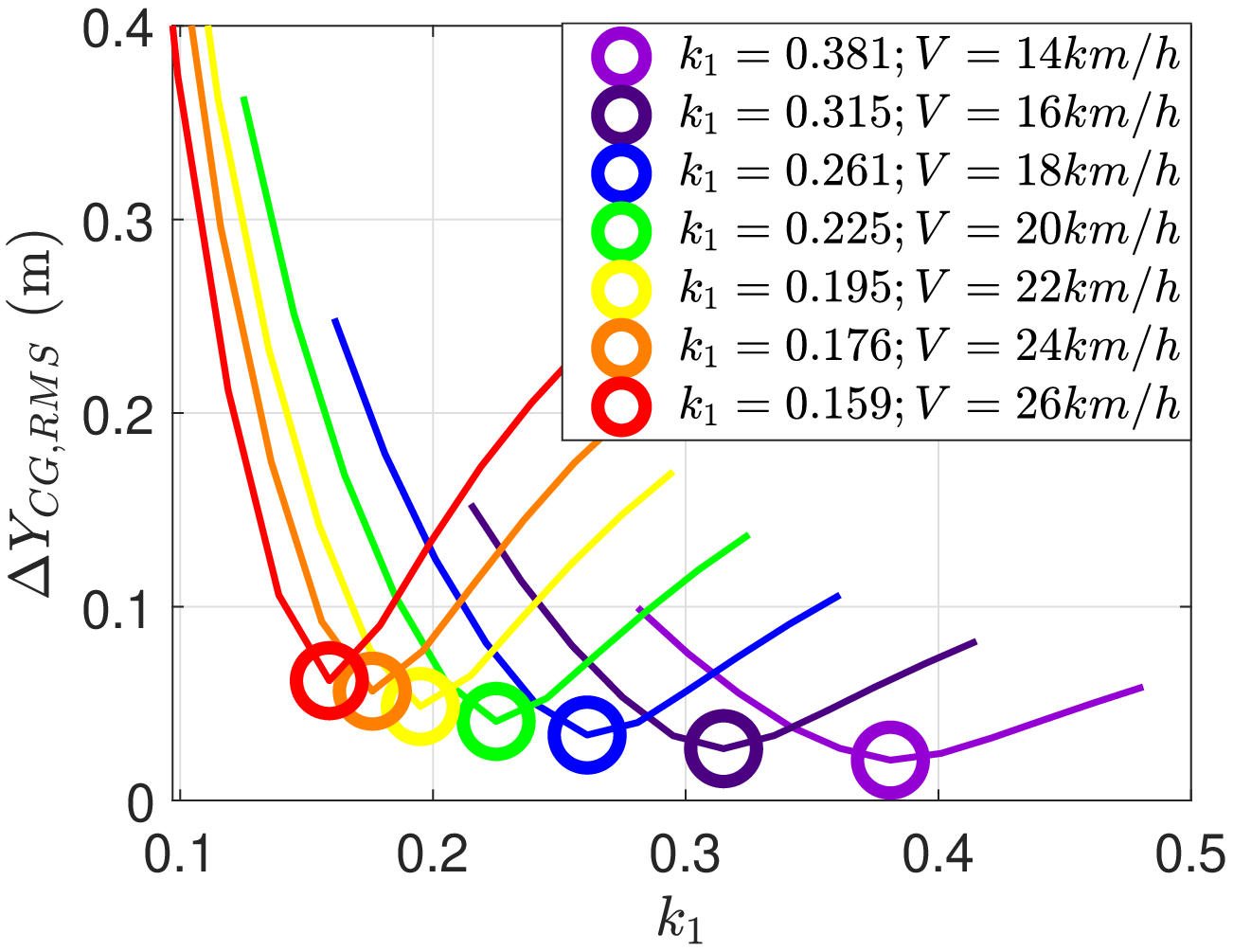}%
\label{fig:03_2lookAheadK1K2}}
\caption{(a) Look-ahead driver model control. (b) Tuning of $k_1$ for the look-ahead driver model keeping $k_2=0.9s$ constant.}
%\label{fig:noiseSets}
\end{figure}

This driver model represents the general control tendency while driving at low-medium lateral accelerations, in which the human operator steers the vehicle to try to align a look-ahead point with the desired trajectory (figure \ref{fig:03_1lookAhead}). Look-ahead driver model based on the cross-track error at the look-ahead point (motivated by \cite{Park1996}) is given by
\begin{align}
\delta &= -k_1\cdot \Delta y_L \label{eq lookaheadDriverModel1}\\
lookahead\: Distance &=k_2\cdot V_x \label{eq lookaheadDriverModel2}\;\;\;\; .
\end{align}
\\
$\delta$ : Steer angle.
\\
$k_1$ : Gain term, a constant for a given vehicle longitudinal speed.
\\
$\Delta y_L$ : Cross-track error of the look-ahead point from the reference trajectory.
\\
$k_2=0.90$ : look-ahead time.

$k_1$ is tuned for various vehicle speeds to have minimum deviation at the CG from the reference trajectory, while driving across region-A of the trajectory shown in figure (\ref{fig:x 06_trajectory}). Observations are presented in figure \ref{fig:03_2lookAheadK1K2}).

$k_1$ is tuned for a constant $k_2$ without considering network delays in the control loop. Although in the presence of delays a human operator can adapt his actions, but keeping $[k_1;\;k_2]$ unchanged ensures no adaptability and highlights performance deterioration due to delays.

\subsubsection{Stanley lateral controller driver, H$_2$}\label{sec2.1.2}
Kinematic Stanley controller \cite{stanley2007} with the reference point at center of front axle, given by
\begin{equation}
\resizebox{\columnwidth}{!}{$
\delta= \begin{cases}\Delta \psi+\tan^{-1} \frac{k \Delta y_F}{V_x} & \text { if }\left|\Delta \psi+\tan^{-1} \frac{k \Delta y_F}{V_x}\right|<\delta_{\max } \\ \:\:\:\delta_{\max } & \text { if } \:\:\Delta \psi+\tan^{-1} \frac{k \Delta y_F}{V_x} \:\: \geq \delta_{\max } \\ -\delta_{\max } & \text { if } \:\: \Delta \psi+\tan^{-1} \frac{k \Delta y_F}{V_x} \leq-\delta_{\max }\end{cases}
$}
\label{eq:stanley}
\end{equation}

\begin{figure}[h]
\centering
    \includegraphics[width=0.9\columnwidth]{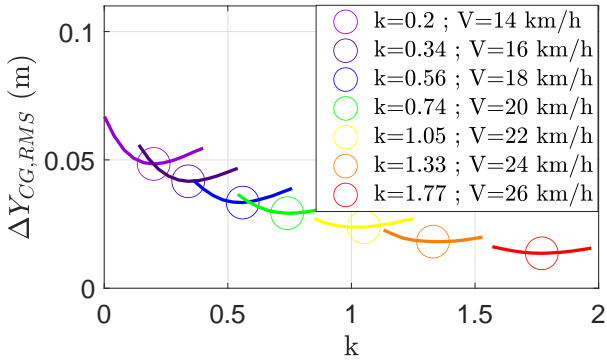}
    \caption{ Tuning of $k$ for the Stanley controller.}
    \label{fig:04_kStanley}%
\end{figure}

$\Delta \psi$ represents the vehicle's heading relative to the nearest segment of the trajectory. The variable $\Delta y_F$ represents the cross-track error at the front axle center. $V_x$ represents the vehicle speed. $k$ is tuned for various vehicle speeds to have minimum deviation at the CG from the reference trajectory while driving across region-A of the trajectory shown in figure (\ref{fig:x 06_trajectory}). Observations are presented in figure \ref{fig:04_kStanley}).
\\

Parameters of both driver models are tuned in only region-A of the trajectory. Region-A is not constant radius but it carries variable curvature across itself as shown in (\ref{fig:x 06_trajectory}). This means the driver models are tuned for variable curvatures.

\subsection{SRPT teleoperation approach with reference-pose decider driver model}\label{sec2.3}
In traditional vehicle teleoperation, where the model-based prediction approach (discussed above) is effective on normal roads and under normal conditions. However, disturbances like strong winds, low-adherence roads, and bumps can alter vehicle dynamics. Parameter estimation techniques presented in articles \cite{Teng, Thomas} can be useful for changes in dynamics that last from medium to high duration. These techniques use sliding window batch estimation, which means that estimations have a lead time. Momentary disturbances can have a significant impact on the vehicle output before the new plant dynamics are estimated at the control station and corrective action is taken by the human operator.

\begin{figure}[h]
\centering
    \includegraphics[width=\columnwidth]{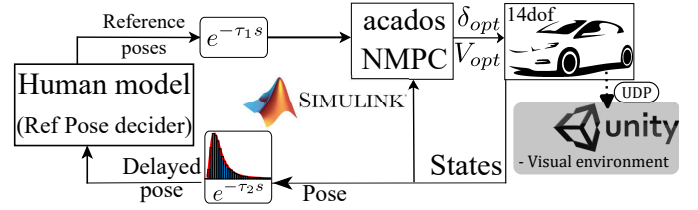}
    \caption{SRPT schematic for vehicle teleoperation simulation. Unity has no role in simulation, it is just to display the manoeuvres.}
    \label{fig:02_srptBlockDiagram}%
\end{figure}

The SRPT approach for vehicle teleoperation differs from traditional methods, in SRPT the human operator transmits reference poses instead of steer-throttle commands to the remote vehicle. These reference poses are generated with a look-ahead time of $[1+\tau_1+\tau_2]s$, which results in the vehicle receiving reference-poses approximately $1s$ ahead of its current position. This horizon of $1s$ is chosen arbitrarily based on the fact that a driver typically steers a vehicle based on upcoming vehicle position. The same time horizon of $\Delta t_{Horizon}=1s$ is also used for the NMPC block to optimize for vehicle steer-speed commands. While the SRPT approach is effective, it represents a departure from conventional vehicle teleoperation, which relies on human operators transmitting steer-throttle commands to the remote vehicle.

\subsubsection{Reference-pose decider driver model}\label{sec2.2.1}
\begin{figure}[h]
%\centering
    \includegraphics[width=1\columnwidth]{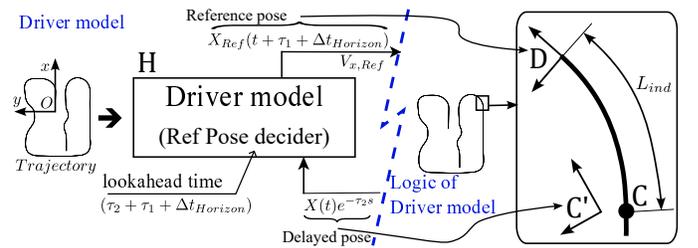}
    \caption{Working principle of the reference-pose decider block. Its task is to choose the future reference pose based on the received vehicle pose and look-ahead distance.}
    \label{fig:05_humanModelSRPT}%
\end{figure}

The task of the human model block is to transmit information that informs the vehicle about its aiming direction. Referring to figure \ref{fig:05_humanModelSRPT}, human model block receives delayed vehicle states, $X(t)e^{-\tau_2 s}$, which consists of vehicle pose, ${P}^{C'}_O$. It is the delayed vehicle pose in global reference frame, $O$. Being aware of the whole trajectory, the human model block first finds the closest point $C$ on the reference trajectory. Then it finds the point $D$, which is $L_{ind}$ distance ahead of point $C$. The $L_{ind}$ is the look-ahead distance govern by below relation:
\begin{equation}
L_{ind} = V_x\cdot\tau + \max(V_x\cdot \Delta t_{Horizon}, \;\;l_F) \label{eq lookaheaDistance}\;\;\;\; .
\end{equation} 

It is lower bounded by $l_F$, the front axle distance from CG. It is linearly proportional to the round trip delay ($\tau=\tau_1+\tau_2$) and to the vehicle speed ($V_x$). The first term tries to compensate for round-trip delay, and the second term aims to generate the terminal condition for the NMPC horizon.

%\color{blue}{
There are two ways in which the aiming direction can be transmitted to the vehicle:
\begin{enumerate}
%\color{blue}
\item Transmit the relative reference pose, ${P}^{D}_{C'}$. It is the relative position and heading of pose-$D$ with respect to pose-$C'$, it acts as a correction term, which tries to bring the vehicle close to the desired trajectory. Upon receive of this relative pose, the vehicle first estimates how much it has already traveled during the round-trip delay and how much more it has to travel. This estimation is possible, as messages are timestamped.

\item Transmit the global reference pose. Transmit the reference pose, $X_{Ref}$, in global reference frame.
\begin{equation}
X_{Ref}={P}^{C'}_O + {P}^{D}_{C'} = {P}^{D}_{O} \label{eq humanPerfectStates}\;\;\;\; .
\end{equation} 

\end{enumerate}
%\textcolor{blue}{
For this paper, we adopt the second approach, which does not require the vehicle to explicitly estimate how much it has travelled during the round-trip delay.
%Since the focus of this paper is the effect of vehicle state estimation error, which is accounted in the simulation environment, this approach is reasonable. 
%}
%The reference pose to be transmitted by the human model is given by:
%As it can be noticed that the point $C'$ is offset from the desired trajectory, ${P}^{D}_{C'}$ is acting as a correction term, which tries to bring the vehicle close to the desired trajectory.
Modeling this driver model for simulation is straightforward as the entire trajectory is pre-known. However, in human-in-the-loop experiments, an equivalent driver model can be obtained, where the correction, ${P}^{D}_{C'}$, can be decided by the human operator. In our previous work \cite{jai2022_3}, this correction term is getting generated online using a steering joystick, briefly represented by the below relation

\begin{equation}
{X}_{Ref}={P}^{C'}_O + \Delta P_{Joystick} \label{eq humanPredictedStates}\;\;\;\; .
\end{equation} 

%${P}^{C'}_O$ - It denotes the predicted vehicle pose, readily available in ${X}\left(t \right) e^{-\tau_2 s} $
%\\
$\Delta P_{Joystick}$ - It is the correction term generated by the augmented lookahead vehicle (blue) outline on the visual interface (figure \ref{fig:00_abstractImgPPT}) with help of joystick steering.
\\

\subsection{NMPC block}
The NMPC block on the vehicle side takes into account the reference poses received, it analyzes the current states of the vehicle, and actuator constraints to generate optimized steer and speed commands. The prediction model of NMPC is presented in our previous work \cite{jai2022_3, jai2022_2}. The objective is to synchronize the target reference pose with the trajectory of the vehicle while minimizing inputs (steer-rate and vehicle acceleration) and maintaining a speed close to the reference speed ($V_{Ref}$) asked by the human operator. It also respects input constraints. One input constraint is the maximum steer-rate of $360^{\circ}/s$, which is due to the actuator constraint of the motor for the steering actuation. Another input constraint is vehicle acceleration and deceleration limits. Further description of NMPC block is presented in the previous works mentioned earlier. A prediction horizon ($\Delta t_{Horizon}$) of 1 second is used, divided into 50 intervals through discrete multiple shooting, and solved by sequential quadratic programming with the real-time NMPC solver ACADOS \cite{Verschueren2018, Verschueren2021}.
\section{Simulation Platform}\label{sec3}
\noindent A faster than real-time simulation test platform for vehicle teleoperation with network delay is developed using Simulink + Unity3D, shown in figure (\ref{fig:smith_1},\ref{fig:02_srptBlockDiagram}). Unity3D is used only to provide visuals of vehicle maneuvers.

Table \ref{tab:vehicleParameters} provides a brief description of the vehicle type used in the 14-dof Simulink vehicle model, which represents a typical FWD passenger vehicle. Table \ref{tab:blockDescriptions} provides additional descriptions of each block, including their working rate.

The $e^{-\tau_2 s}$, Human model, and $e^{-\tau_1 s}$ blocks work synchronously with each other at $30\:Hz$ to simulate the usual discrete nature of video streaming to the control station. The downlink delay ($\tau_2$) is considered a variable delay to simulate usual network delays, while the uplink delay ($\tau_1$) is considered a constant of $0.060s$ due to its lower magnitude and variability. To simulate the downlink delay, a generalized extreme value distribution, $GEV(\xi=0.29, \mu_{GEV}=0.200, \sigma=0.009)$ is used~\cite{Zheng2020, jai2022}. Positive $\xi$ means that the distribution has a lower bound of $(\mu_{GEV}-\frac{\sigma}{\xi})\approx0.169$~s$\:(>0)$ and a continuous right tail based on extreme value theory, keeping the variable downlink delay in the range of $0.169s-0.300 s$.

\begin{table}[ht]
\centering
\caption{14-dof model: Vehicle brief characteristics.}
\label{tab:vehicleParameters}
\begin{tabular}{cc}
\hline
\textbf{Parameter}                & \textbf{Value}             \\ \hline
$m$                               & 1681\:kg                   \\ \hline
$I_z$                             & 2600\:kg\,s$^2$            \\ \hline
$[m_F;\:m_R]$                     & [871.6;\:809.4]\:kg        \\ \hline
$[l_F;\:l_R]$                     & [1.3;\:1.4]\:m             \\ \hline
\end{tabular}
\end{table}

\begin{table}[h]
\centering
\caption{Description of the blocks used in the simulation platform.}
\begin{tabular}{|l|l|l|}
\hline
\multicolumn{1}{|c|}{\textbf{Block}} & \multicolumn{1}{c|}{\textbf{Description}}                                                            & \multicolumn{1}{c|}{\textbf{Rate}} \\ \hline
Vehicle 14dof                        & \begin{tabular}[c]{@{}l@{}}A 14dof vehicle model to simulate\\ a real vehicle\end{tabular}           & 1000 Hz                            \\ \hline
$e^{-\tau_2 s}$                      & Variable network downlink delay                                                                      & 30 Hz                              \\ \hline
Human model                          & Ref Pose decider driver                                                                              & 30 Hz                              \\ \hline
$e^{-\tau_1 s}$                      & \begin{tabular}[c]{@{}l@{}}Constant network uplink delay\\ $\tau_1 = 0.060s$\end{tabular}            & 30 Hz                              \\ \hline
NMPC                                 & \begin{tabular}[c]{@{}l@{}}Non-linear model predictive controller\\ Acados toolkit\end{tabular}      & 50 Hz                              \\ \hline
Unity                                & \begin{tabular}[c]{@{}l@{}}An external block, to visualize the real\\ vehicle maneuvers\end{tabular} & 100 Hz                             \\ \hline
\end{tabular}
\label{tab:blockDescriptions}
\end{table}
\section{Experimental Setup}\label{sec4}
\begin{figure}[h]
    \centering
    \includegraphics[width=1\columnwidth]{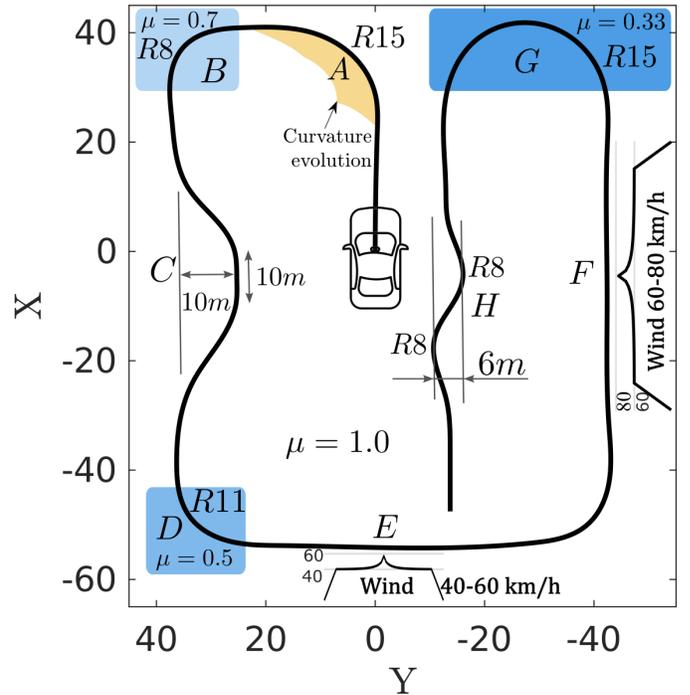}
    \caption{Track contains various sections A-H of difficult manoeuvres and worst-case environmental conditions.}
    \label{fig:x 06_trajectory}
\end{figure}
\noindent Figure \ref{fig:x 06_trajectory} shows a 438m test track consisting of eight regions labeled from A to H. These regions simulate increasingly challenging maneuvers and severe environmental conditions. Region A involves cornering with a radius of $15$m ($R15$), B involves cornering ($R8$) on a surface with road adherence coefficient of $\mu=0.7$. Region C is double lane change, D involves cornering with $\mu=0.5$, E-F includes strong lateral wind with a Chinese hat profile \cite{Baker2001, Baker2022}, G involves a U-turn with $\mu=0.33$, and H involves a slalom. All the curves have gradually changing curvature, as shown for region-A. The objective is to follow the track centerline as closely as possible, with a maximum vehicle speed limit of $V_{Ref}$, specified by the human block. It is anticipated that during difficult manoeuvres, the NMPC block regulates the vehicle speed ($V_{opt}$) to minimize the cross-track error, which is a desirable behavior.

To compare SRPT performance over Smith-predictor performance, a total of eight modes are considered (as given below):
%\begin{enumerate}[label=]
%\begin{enumerate}
%\begin{enumerate}[1.]
\begin{enumerate}[label=\arabic*.]
    \item NoDelay\,-LookAhead driver
    \item Delay\ \ \ \ -LookAhead driver
    \item Delay\ \ \ \ -LookAhead driver (Smith)
    \item NoDelay - Stanley driver
    \item Delay\ \ \ \ \,- Stanley driver
    \item Delay\ \ \ \ \,- Stanley driver (Smith)
    \item NoDelay - RefPoses driver (SRPT)
    \item Delay\ \ \ \ \,- RefPoses driver (SRPT)
\end{enumerate}
Also, to assess performances over a range of vehicle speeds, each mode is tested on vehicle speeds ranging from $V_{Ref} = 14$ km/h to $V_{Ref} = 26$ km/h in succession.
\section{Results and discussion}\label{sec5}
\noindent The extent of performance degradation due to delays is expected to vary based on the vehicle speed and maneuver promptness. At lower speeds, latency effects are less pronounced as the vehicle has more time to respond to commands.

As speed increases, the available response time to perform a maneuver decreases, and latency can significantly impact the accuracy and safety of the maneuver. The chosen test track has an increasing level of difficulty along its length and will be traversed at various speeds, one at a time, for this study. Just to understand the approximate steer-rate requirement for the track at corresponding vehicle speeds, the Ackermann steering relation can be used as given below:

\begin{align}
\delta(s) &= \tan^{-1}\left[\frac{L}{R(s)} \right] \label{eq ackermannSteer}\\
steer\:rate, \frac{d \delta}{dt}(s) &= \frac{d \delta}{ds} \cdot \frac{ds}{dt} = \frac{d \delta}{ds} \cdot V\label{eq steerRateRequirement}
\end{align}
\\
$\delta(s)$  \ : Steer angle.
\\
$L$ \ \ \ \ : Wheelbase.
\\
$R(s)$ : Radius of curvature along the track length ($s$).
\\
$V$ \ \ \ : Vehicle speed.

\begin{figure}[h]
\centering
    \includegraphics[width=1\columnwidth]{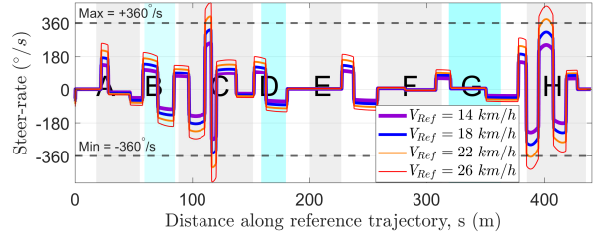}
    \caption{Approximate steer-rate requirement for the track for various vehicle speeds.}
    \label{fig:10_steerRateRequirement}%
\end{figure}

In Figure \ref{fig:10_steerRateRequirement}, the steer-rate requirement for different vehicle speeds along the track is shown. It can be observed that in regions C and H, the required steer-rate exceeds the maximum steer-rate constraint beyond the reference speed of $22$ km/h. This implies that at higher speeds, the vehicle may not be able to perform the necessary steering maneuvers, which can lead to a higher cross-track error and reduced performance. In addition to the steer-rate constraint, the track includes other factors that can negatively impact performance. Hereafter, the RMS of the cross-track error at the vehicle CG will be used as the primary performance index to compare the performance of the respective vehicle teleoperation modes.

\begin{figure}[h]
\centering
    \includegraphics[width=1.0\columnwidth]{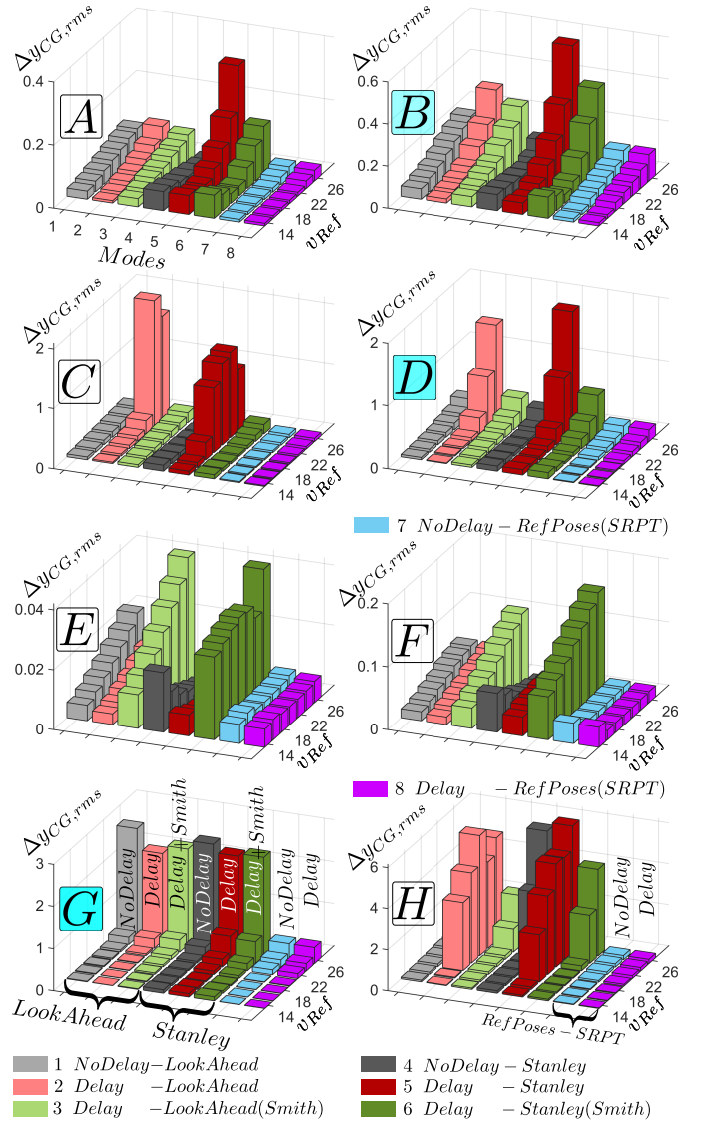}
    \caption{Vehicle teleoperation simulation result with various modes at $V_{Ref}=26\;km/h$. SRPT vehicle teleoperation accurately traces the track, even in the presence of variable delays.}
    \label{fig:08_crossTrackError3dBars_edited}%
\end{figure}

Figure \ref{fig:08_crossTrackError3dBars_edited} presents a quantitative analysis of cross-track errors observed in regions A-H for different vehicle speeds and teleoperation modes. In region A-D, the Smith predictor ameliorates the negative effect of delays and tries to reduce the cross-track error to its respective undelayed mode (the green bars are shorter than the red bars). However, the SRPT mode, even with delay (purple bars), resulted in significantly smaller cross-track errors. In regions E-F with strong crosswinds, the Smith predictor approach results in larger cross-track errors because it is unaware of the wind disturbances. In contrast, the NMPC controller in the SRPT mode takes vehicle states as input, leading to a significant improvement in teleoperation. In region-G ($\mu=0.33$), for the high-speed lap, all teleoperation modes except the SRPT mode resulted in high lateral slip and therefore high cross-track errors. In region-H (the slalom), the cumulative impact of both the steer-rate constraint and delay in the control loop deteriorates the performance. Even in this region, the SRPT mode demonstrated a significant reduction in cross-track error.

\begin{figure}[h]
\centering
    \includegraphics[width=1.0\columnwidth]{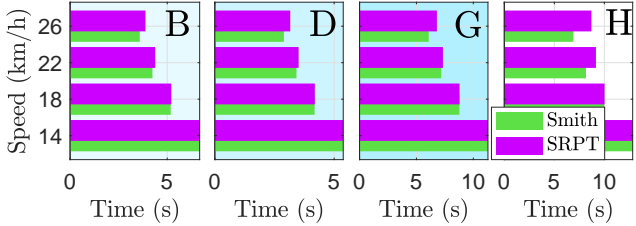}
    \caption{Completion time comparison SRPT vs Smith mode for [B,D,G,H] regions.}
    \label{fig:11_completionTime}%
\end{figure}

\begin{figure}[h]
\centering
    \includegraphics[width=1.0\columnwidth]{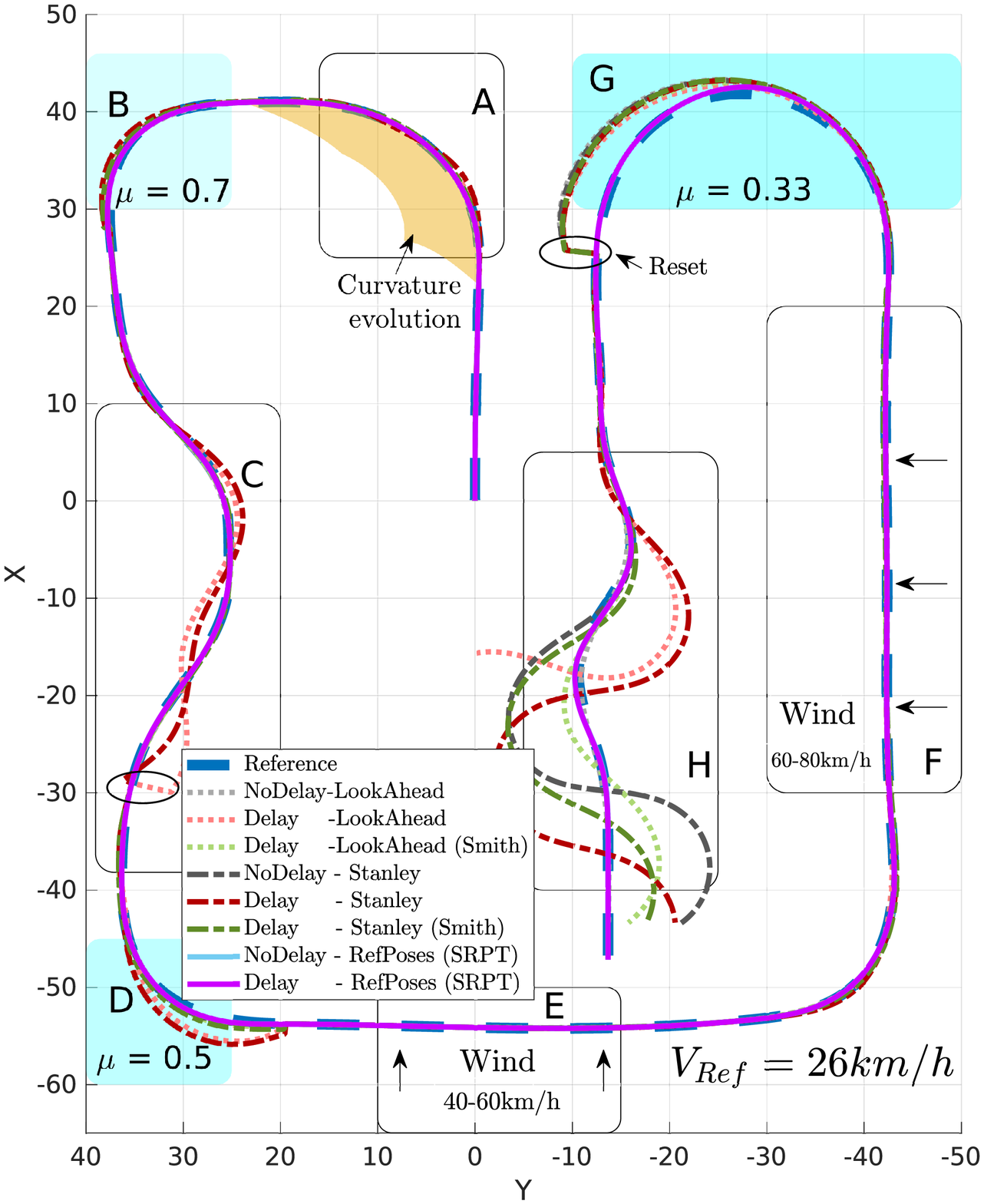}
    \caption{Vehicle teleoperation simulation result with various modes at $V_{Ref}=26\;km/h$. SRPT vehicle teleoperation accurately traces the track, even in the presence of variable delays.}
    \label{fig:06_trajectoryTraversed}%
\end{figure}

The analysis of the results revealed that the primary reason for the superior tracking performance of the SRPT mode is its ability to moderate the vehicle speed appropriately in areas where it is necessary to minimize the cross-track error. Consequently, this leads to a slight increase in the completion time, as shown in Figure \ref{fig:11_completionTime}. An example of the trade-off between completion time and safety can be seen in region-H (slalom) where SRPT mode resulted in a $25\%$ increase in completion time at $V_{Ref} = 26\; km/h$. Despite the longer completion time, this mode ensures higher safety and minimizes cross-track error, which is particularly important for this tight slalom at this vehicle speed.

Figure \ref{fig:06_trajectoryTraversed} presents the trajectory traversed with all the teleoperation modes for $V_{Ref}=26\;km/h$. It qualitatively shows better performance of SRPT approach even in the presence of all the disturbances and variable delays. The red trajectory of look-ahead driver model resulted in big oscillations due to network delay and due to steer-rate saturation. If any mode deviates significantly from the track, to the extent that it may compromise the results of the subsequent region, the mode is reset before entering the new region, while maintaining the vehicle's initial speed as the reference speed.

These large deviations and oscillations are not present in SRPT approach because NMPC block accounts for the steer-rate limitation and subsequently decelerates the vehicle to allow for more time to steer.

\begin{figure}[h]
\centering
    \includegraphics[width=1.0\columnwidth]{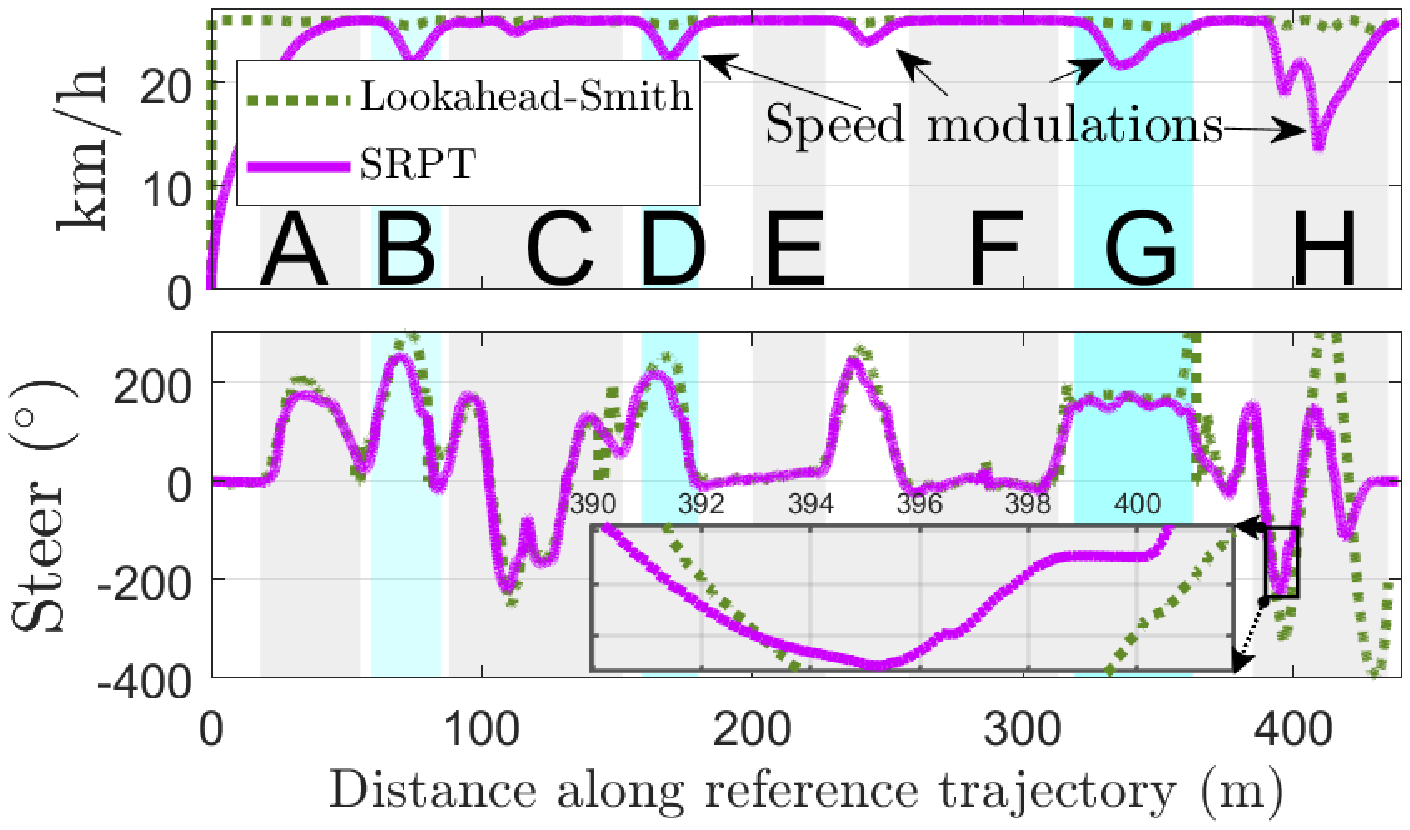}
    \caption{Vehicle teleoperation simulation result with various modes at $V_{Ref}=26\;km/h$. SRPT vehicle teleoperation accurately traces the track, even in the presence of variable delays.}
    \label{fig:09_speedSteerModulations}%
\end{figure}

Figure \ref{fig:09_speedSteerModulations} shows the vehicle speed profile along the track length for Lookahead-Smith mode and SRPT mode, both in the presence of network delays.  The SRPT mode implemented automatic speed modulations, which were noticeable in all cornering regions, particularly in the slalom region. These modulations helped steer in advance, as shown in the zoomed rectangle inside the figure.

Interestingly, the performance of the SRPT mode, with and without network delay, is similar (blue bars and purple bars are similar in height in figure \ref{fig:08_crossTrackError3dBars_edited}). This can be attributed to the difference in SRPT mode operating principle, wherein the vehicle receives reference poses instead of steer commands from the control station.
\section{Conclusion}\label{sec6}

In this paper, the novel SRPT approach for vehicle teleoperation is tested, where the remote vehicle receives reference poses instead of steer commands. A simulation framework was set up using a Simulink environment to test the approach under variable network delays (250-350ms). The performance of SRPT was compared with the Smith predictor approach, accounting for two driver models: the Lookahead and Stanley models. The simulation experiments included various modes and vehicle speeds ($V_{Ref} = 14-26\;km/h$), and the performance index is the rms of the cross-track error in various regions of the test track.

The results showed that the SRPT approach is effective for all maneuvers and environmental disturbances at all vehicle speeds considered in the experiment. It resulted in significantly lower cross-track error compared to the other teleoperation modes. In low adhesion road and slalom regions, SRPT mode showed significant improvement in path tracking performance compared to other modes. The SRPT mode automatically moderates vehicle speed at the instant when it was necessary, which allowed more time to steer for a maneuver. This led to a slight increase in completion time in difficult maneuvers.

Due to its different working principle, the SRPT approach performance does not deteriorate with the presence of network delays. However, implementing the SRPT approach in real-world teleoperation would require a state estimator. Therefore, designing a state estimator and investigating the effect of its estimation inaccuracy on the performance of the SRPT approach are left for future research. Eventually, this new approach is planned to be deployed in a real vehicle and control-station with an aim to perform real-world vehicle teleoperation experiments.

\bibliographystyle{IEEEtran}
\bibliography{sample}

\end{document}